\documentclass[journal]{IEEEtran}
\usepackage{amsmath,amsfonts}
\pdfoutput=1
\usepackage{amssymb}
\usepackage{algorithmic}
\usepackage{array}
\usepackage[caption=false,font=normalsize,labelfont=sf,textfont=sf]{subfig}
\usepackage{textcomp}
\usepackage{stfloats}
\usepackage{url}
\usepackage{verbatim}
\usepackage{graphicx}
\def\BibTeX{{\rm B\kern-.05em{\sc i\kern-.025em b}\kern-.08em
    T\kern-.1667em\lower.7ex\hbox{E}\kern-.125emX}}
\usepackage{balance}
\usepackage{algorithmic}
\usepackage{algorithm}
\usepackage{multirow}

\begin{document}
\title{GCFL: A Gradient Correction-based Federated Learning Framework
for Privacy-preserving CPSS}
\author{Jiayi Wan, Xiang Zhu*, Fanzhen Liu, Wei Fan, Xiaolong Xu
\thanks{Jiayi Wan is with the School of Software Engineering, Nanjing University of Information Science and Technology, Nanjing, China (email: wanjiayi117@gmail.com).\par
% Guangming Cui is with Jiangsu Province Engineering Research Center of Advanced Computing and Intelligent Services, School of Software, Nanjing University of Information Science \& Technology, P.R. China, and State Key Laboratory for Novel Software Technology, Nanjing University, P.R. China. (email: gcui@nuist.edu.cn).\par
Xiang Zhu is with the College of Meteorology and Oceanography, National University of Defense Technology, Changsha, China (email: zhuxiang@nudt.edu.cn).
Fanzhen Liu is with CSIRO’s Data61, Sydney, Australia (email: Fanzhen.Liu@data61.csiro.au).\par
Wei Fan is with the Medical Sciences Division, University of Oxford, Oxford, UK (email: frankfanwei@outlook.com).\par
% Muhammad Bilal is with the School of Computing and Communications, Lancaster University, Lancaster, UK (email: m.bilal@ieee.org).\par
Xiaolong Xu is with Jiangsu Province Engineering Research Center of Advanced Computing and Intelligent Services, and Jiangsu Collaborative Innovation Center of Atmospheric Environment and Equipment Technology, Nanjing University of Information Science and Technology, Nanjing, China, State Key Laboratory for Novel Software Technology, Nanjing University, Nanjing, China (email: xlxu@ieee.org). \par
(Corresponding author: Xiang Zhu.)

}}

\markboth{
% Journal of \LaTeX\ Class Files,~Vol.~18, No.~9, September~2020}%
}
{
% How to Use the IEEEtran \LaTeX \ Templates
}

\maketitle

\begin{abstract}
Federated learning, as a distributed architecture, shows great promise for applications in Cyber-Physical-Social Systems (CPSS). In order to mitigate the privacy risks inherent in CPSS, the integration of differential privacy with federated learning has attracted considerable attention. Existing research mainly focuses on dynamically adjusting the noise added or discarding certain gradients to mitigate the noise introduced by differential privacy. However, these approaches fail to remove the noise that hinders convergence and correct the gradients affected by the noise, which significantly reduces the accuracy of model classification. To overcome these challenges, this paper proposes a novel framework for differentially private federated learning that balances rigorous privacy guarantees with accuracy by introducing a server-side gradient correction mechanism. Specifically, after clients perform gradient clipping and noise perturbation, our framework detects deviations in the noisy local gradients and employs a projection mechanism to correct them, mitigating the negative impact of noise. Simultaneously, gradient projection promotes the alignment of gradients from different clients and guides the model towards convergence to a global optimum.  We evaluate our framework on several benchmark datasets, and the experimental results demonstrate that it achieves state-of-the-art performance under the same privacy budget.
\end{abstract}


\section{Introduction}
\IEEEPARstart{T}{he} proliferation of Cyber-Physical Systems (CPSS) has revolutionized a wide range of sectors, including smart manufacturing, autonomous driving \cite{Zhao2023}, and healthcare \cite{zzz7, Sun2023}, driving substantial advancements in both efficiency and intelligence \cite{xu2025c2lrec}. By seamlessly integrating computation, communication, and physical processes, CPSS enables sophisticated real-time monitoring and control within complex systems, offering significant improvements in operational performance and decision-making processes \cite{Ding2024}. However, the expanding scope of CPSS applications and the rapid pace of technological advancements present dual challenges: a dramatic increase in data processing demands and escalating system complexity \cite{yao2023differential}.  Traditional centralized computing architectures are increasingly inadequate to address these demands, particularly when confronted with large-scale data processing and stringent real-time requirements \cite{zzz8,xu2024xrl}.\par

Furthermore, CPSS relies heavily on sensors and devices for real-time data acquisition and decision support, involving the processing of vast amounts of sensitive information \cite{zhou2025decentralized, Kanthuru2023}. Ensuring data privacy while maintaining system performance has become a critical concern. Federated Learning (FL) \cite{zzz9} offers a promising solution as a novel distributed learning paradigm. By distributing data processing tasks across multiple nodes for parallel execution, FL significantly enhances computational efficiency, fault tolerance, and real-time responsiveness in CPSS. Moreover, FL enables collaborative model training without exchanging raw data, mitigating the traditional data silo problem and facilitating knowledge sharing among participants, thereby improving the performance of various tasks within CPSS.\par

While FL offers significant advantages in distributed model training, recent studies have exposed vulnerabilities in FL systems that exclusively rely on the exchange of model parameters \cite{Li2025,10475968}. Attackers can infer sensitive information \cite{song2017machine} or even reconstruct individual data \cite{mahmood2018pairing,zzz1} by analyzing model updates or gradients, even without direct access to the training data. Such risks present substantial threats to the privacy of CPSS. Moreover, more sophisticated attacks, such as linkage attacks, can re-identify anonymized data, further compromising individual privacy \cite{melis2019exploiting}. As a result, ensuring data privacy, particularly preventing the leakage of training data, becomes paramount in the context of CPSS.\par

Differential Privacy (DP) \cite{Tian2024} has emerged as a crucial tool for safeguarding data privacy in FL. By adding noise to data, DP ensures that individual information cannot be gleaned from analyzing model outputs. The advent of Differentially Private Stochastic Gradient Descent (DP-SGD) has enabled the effective application of DP to deep learning training. Through gradient clipping and noise injection, DP-SGD effectively protects the privacy of training data, preventing attackers from inferring sensitive individual information from model outputs. Within the FL framework, local clients utilize DP to add noise to model parameters or gradients, masking the influence of individual data. Clients then upload the updated models to a central server, which aggregates the updates from multiple clients to generate a globally improved model with enhanced learning and generalization capabilities.\par

While recent research demonstrates the efficacy of combining DP and FL for data privacy, the interplay between privacy preservation and distributed training often significantly impacts model performance, especially in resource-constrained CPSS environments. To address this challenge, researchers have proposed various improved federated aggregation algorithms. For example, adaptively reducing the noise scale can balance privacy budgets and model performance, offering new avenues for privacy preservation and performance optimization in CPSS \cite{fu2022adap}. Furthermore, recent efforts have explored evaluating gradients based on validation tests in each iteration, applying only convergence-inducing updates to the model. However, these approaches primarily focus on reducing the added noise during the iterative process or discarding heavily noise-affected gradients, without explicitly removing convergence-hindering noise or correcting noise-affected gradients during gradient aggregation. This often leads to suboptimal performance and slow convergence under high noise levels. Therefore, precisely controlling noise and mitigating its negative impact on the model remains a crucial challenge for enhancing CPSS performance in edge computing contexts.\par

Inspired by the aforementioned observations, this paper proposes a novel DP-FL framework that provides strong privacy guarantees through client-side gradient perturbation and employs an innovative server-side gradient correction mechanism to explicitly mitigate the adverse effects of noise on model convergence, ensuring the accuracy and effectiveness of deep learning models trained on sensitive personal data.  Specifically, we evaluate gradients from different clients at the server and project those gradients whose noise contributions lead to model divergence. This projection operation effectively removes noise that disrupts the correct update direction of model parameters, maintaining stable model performance even under high noise levels.  Simultaneously, the projection brings gradients from different clients closer, correcting the model update direction from local optima towards the global optimum. This local optima correction mechanism enhances the performance of DP-FL in CPSS while guaranteeing strong privacy for client data. Our main contributions are summarized as follows:
\par

\begin{itemize}
    \item  We propose a novel DP-FL framework designed to optimize privacy preservation in CPSS while simultaneously enhancing model performance. This framework ensures client data privacy in high-noise edge environments while improving model performance.
    \item We introduce an innovative gradient correction mechanism. This mechanism leverages gradient projection to mitigate the adverse effects of noise on model updates, maintaining stable model performance even under high noise levels. Furthermore, gradient projection promotes alignment of gradients from different clients, guiding model convergence towards a direction closer to the global optimum.
    \item Extensive experiments are conducted on three benchmark datasets. State-of-the-art performance is demonstrated compared to mainstream baseline models under the same privacy budget.
\end{itemize}

\begin{figure*}[htpb]
    \centering
    \includegraphics[width=0.9\linewidth]{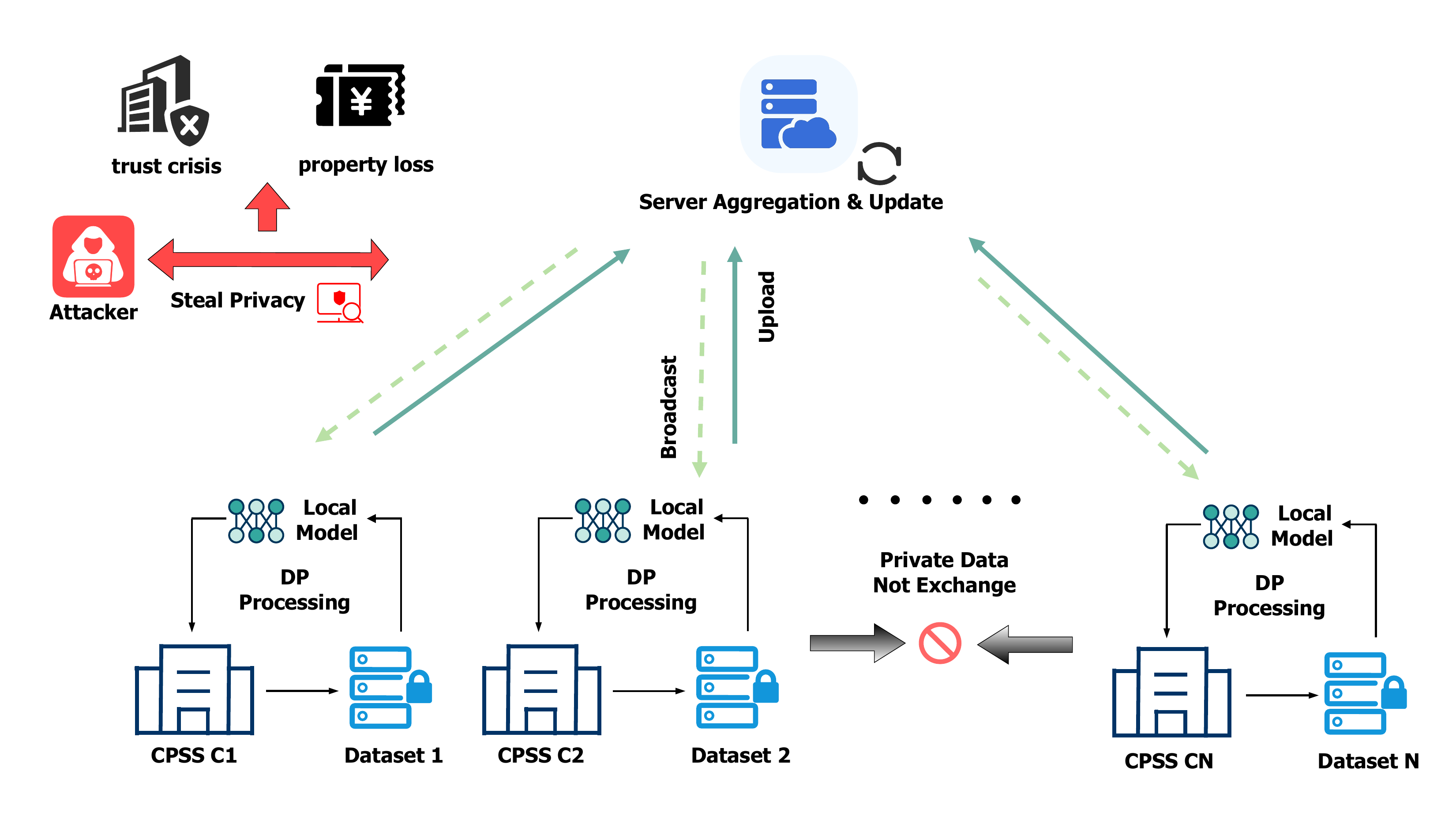}
    \caption{An application of differential privacy and federated learning involves multiple clients training models locally, leveraging differential privacy to protect data privacy. A central server aggregates updates from multiple clients to produce a global model with enhanced learning and generalization capabilities.
}
    \label{fig:figure_1}
    \vspace{+0.2cm}
\end{figure*}

\section{Related work}

The proliferation of CPSS in industrial applications hinges on effective data sharing and collaboration \cite{zzz2}. However, growing concerns surrounding data privacy pose significant challenges to traditional data sharing paradigms.  This challenge is further exacerbated in edge computing environments, where distributed data storage and processing amplify the risk of privacy breaches \cite{zzz3}.  To enable multi-client data training without compromising individual privacy, researchers have explored integrating techniques such as differential privacy, homomorphic encryption, and secure multi-party computation into federated learning \cite{munjal2023systematic}.  Our work focuses on the synergy between differential privacy and federated learning to address these privacy concerns, specifically aiming to optimize the trade-off between model performance and privacy preservation within the context of edge computing. \par

\textbf{Federated learning.} FL enables multiple edge devices to collaboratively train deep learning models under the control of a central server. Clients train on local datasets and send only model updates, rather than raw data. The central server aggregates these updates iteratively until a set number of training rounds is completed or a performance threshold is achieved. Initially proposed by Google, FL gained traction with the FedAvg algorithm introduced by McMahan et al. \cite{mcmahan2017communication}, which employs weighted averaging of client updates on the server. Subsequent research has addressed two key challenges in FL: $1)$ Data Heterogeneity. The non-independent and identically distributed nature of data across clients poses a significant hurdle. To mitigate this, Li et al. \cite{li2020federated} introduced regularization terms in local model updates to reduce deviations from the global model. Karimireddy et al. \cite{karimireddy2019scaffold} utilized variance reduction approaches to counteract client drift in the process of local updates. Further, to enhance privacy and reduce the computational burden, Dai et al. \cite{10.1145/3702995} integrated FL with an isolation forest. Xu et al. \cite{xu2025blockchain} introduced a distributed privacy-preserving framework. Adaptive aggregation mechanisms have also been explored for global model updates. To address the slow convergence often observed in FL with complex models or high data heterogeneity, Jhunjhunwala et al. \cite{jhunjhunwala2023fedexpspeedingfederatedaveraging} devised a FL optimization method with dynamic server step size adjustment applied to pseudo-gradients. $2)$ Robust Privacy Guarantees. Even with decentralized datasets in FL, user privacy remains vulnerable. A potentially "honest-but-curious" server or third-party access to exchanged model parameters during or after training can expose sensitive information \cite{fredrikson2015model, geiping2020inverting}. DP, achieved by adding random noise to model updates, has become a popular approach for integration with FL. Our work aims to address both the local drift issue in model updates and provide robust privacy guarantees.\par

\textbf{Differential Privacy.} Differential Privacy (DP) is primarily a privacy‑preserving technique for data analysis, underpinned by a rigorous mathematical framework for quantifying and enforcing data privacy. It was originally developed for use on data‑collection servers. It prevents differential attacks by introducing precisely calibrated noise to statistical query outputs, making it nearly impossible for an attacker to distinguish between two adjacent datasets.  Dwork pioneered the original definition of differential privacy and introduced the Laplace mechanism as a key enabling technique \cite{dwork2006differential}.  Subsequently, relaxed versions of DP were proposed to improve data utility and model performance while still offering meaningful privacy protection \cite{shen2023data}.  For instance, Dong et al. explored the use of Gaussian noise in differential privacy as an alternative to the traditional Laplace mechanism \cite{dong2022gaussian}.  Building on Rényi divergence, Mironov et al. proposed Rényi Differential Privacy \cite{mironov2017renyi}.  Differentially Private Stochastic Gradient Descent (DP-SGD) marked a significant step by integrating differential privacy with deep learning.  It employs Poisson sampling to select a batch of samples randomly, computes and clips each sample's gradient, and then adds Gaussian noise to each gradient based on differential privacy principles, preventing unintentional leakage of private training data. While numerous efforts have focused on practical applications of DP in industry, such as Dankar et al.'s exploration of DP in healthcare data protection \cite{dankar2013practicing} and Hernandez et al.'s investigation of optimal local DP mechanisms for varying privacy needs \cite{hernandez2024comparative}, these often face a trade-off between strict privacy budgets and high performance.  Recently, Fu et al. \cite{fu2023dpsuracceleratingdifferentiallyprivate} achieved notable results by selectively applying beneficial gradient updates based on a validation testing strategy, discarding noisy or unhelpful updates that hinder convergence in traditional DP-SGD.  Furthermore, adaptive gradient clipping methods 
 \cite{fu2022adap} and simulated annealing-based DP schemes \cite{fu2022sa} have been proposed to enhance model performance under high noise conditions. These works share similarities with our approach.
\par

However, we observe that existing work primarily focuses on reducing the magnitude of added noise during the iterative process or discarding heavily noise-affected gradients, without explicitly correcting the negative impact of noise on model updates. This often leads to suboptimal performance and slow convergence, particularly under high noise regimes. We propose GCFL, which incorporates a gradient correction mechanism to achieve data privacy while enabling efficient model training and updates on resource-constrained edge devices.

\section{Preliminaries}
\noindent 
\begin{figure*}[ht]
    \centering
    \includegraphics[width=0.95\linewidth]{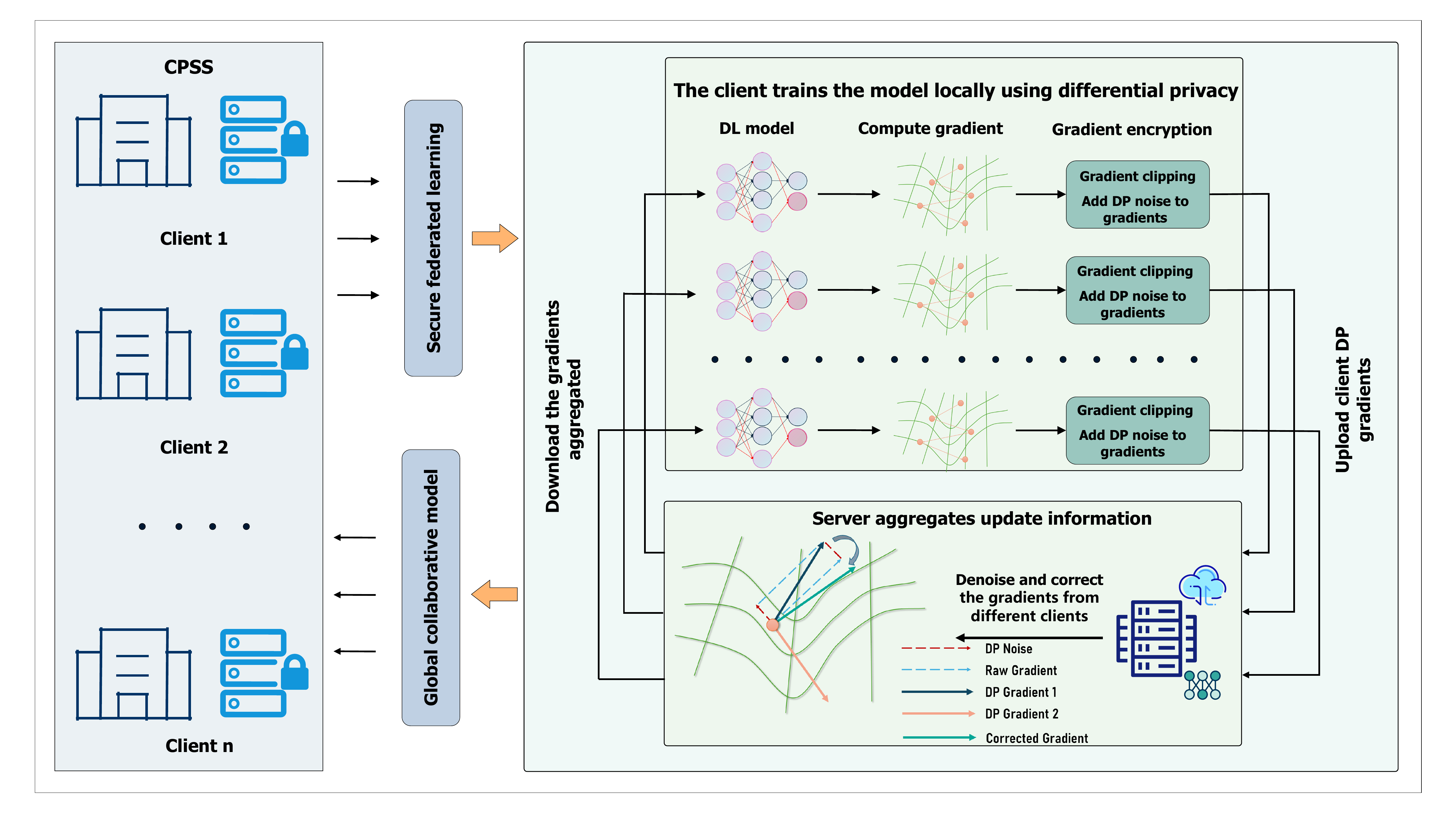}
    \captionsetup{justification=centering}
    \caption{Overview of GCFL: The server receives the noisy model gradients, applies a gradient correction mechanism to adjust gradients that deviate from the global optimal direction, and then returns the aggregated gradients to the client for training.}
    \label{fig:overview}
    \vspace{+0.2cm}
\end{figure*}
% \vspace{-1 cm}
\noindent
We aim to maximize data privacy while applying federated learning to CPSS. Integrating differential privacy has become a standard approach for protecting user privacy \cite{zzz4, zhou2024federated}, as illustrated in Figure \ref{fig:figure_1}.\par 

\textit{Definition 3.1.} (\textbf{Differential Privacy}). The core objective of differential privacy is to ensure that the presence or absence of a single data point in the input dataset does not significantly affect the output statistics \cite{dwork2014algorithmic}. A mechanism satisfies $(\epsilon, \delta)-DP$ if, for any two neighboring datasets $D$ and $D'$ differing by at most one data point, and for any output subset $S \subseteq Range(M)$:
\begin{equation}
    Pr[M(D) \in S] < e^\epsilon Pr[M(D') \in S] + \delta ,
\end{equation}
where $\epsilon$ is the privacy loss parameter, controlling the strength of the privacy guarantee. The parameter $\delta$ represents the probability that the algorithm may fail to satisfy the privacy guarantee. In practice, $\delta$ should be negligibly small.\par

\textit{Definition 3.2.} (\textbf{Rényi Differential Privacy (RDP)}). In the DP-FL framework, the number of iterations required for model convergence is often substantial, necessitating a more concise privacy loss composition measure.  To analyze the cumulative privacy loss more accurately, RDP, based on Rényi divergence \cite{huang2023higher}, extends the traditional definition of differential privacy.  Given two probabilities
distributions $P$ and $Q$, the Rényi divergence is defined as:
\begin{equation}
    D_\alpha(P \parallel Q)= \frac{1}{\alpha-1}lnE_{x\sim Q}[(\frac{P(x)}{Q(x)})^\alpha] ,
\end{equation}
where $\alpha$ is the order of the Rényi divergence, controlling the sensitivity of the divergence. As $\alpha$ approaches 1, the Rényi divergence converges to the classic Kullback-Leibler divergence.\par
RDP measures the privacy difference in the output distributions of a mechanism on adjacent datasets using Rényi divergence. Specifically, a mechanism $M$ satisfies the conditions of $(\alpha, R)-RDP$  if:
\begin{equation}
    D_\alpha(M(D))\parallel M(D')) \leq \epsilon .
\end{equation}
\par
It is important to note that RDP provides an alternative approach for analyzing privacy composition, aiming to offer a tighter analysis of privacy guarantees.\par

\textit{Definition 3.3.} (\textbf{Conversion from RDP to DP \cite{balle2020hypothesis}}). RDP can be converted into standard Differential Privacy. Given an $(\alpha, R)-RDP$ guarantee, the relationship can be expressed as:
\begin{equation}
    \epsilon = R + ln(\frac{(\alpha-1)}{\alpha}) - \frac{(ln\delta + ln\alpha)}{(\alpha-1)} .
\end{equation}
\par In complex medical data protection scenarios, RDP as an extension of traditional DP, offers a more flexible tool for privacy analysis.

\textit{Definition 3.4.} (\textbf{Federated Learning}). FL is a distributed learning framework that has garnered widespread attention in recent years. In the FL framework, multiple client devices collaboratively train a deep learning model under the coordination of a central server, without the need to share raw data that may contain sensitive information. In contrast, clients keep their data locally and contribute to the collective learning process by training models locally and then sending model parameters to the central server for aggregation. Subsequently, the server aggregates the updates of all model parameters and redistributes the improved model to all participants. Formally, the server aggregates the parameters received from N clients as follows:
\begin{equation}
    w = \sum_{i=1}^{N} p_iw_i,
\end{equation}
where $w_i$ presents the model parameters from the $i-th$ client, $w$ represents the model parameters on the server side, $N$ is the number of clients, and $p_i$ is the weight of the model parameters from the i-th client. Therefore, the optimization objective of federated learning can be defined as:
\begin{equation}
    w^* = arg min\sum_{i_1}^N p_iL_i(w,D_i),
\end{equation}
where $L(\cdot)$ is the local loss function used by the $i-th$ client.

\textit{Definition 3.5.} (\textbf{Horizontal Federated Learning (HFL)}). HFL is applicable to scenarios where participating clients hold different samples but share the same feature space. This means that each client's data samples are distinct, but these samples share the same features. The optimization objective of HFL is to train a general global model that is applicable to all clients' data.

\textit{Definition 3.6.} (\textbf{Vertical Federated Learning (VFL)}). VFL is applicable to scenarios where the data is partitioned along the feature space. Each client possesses different features for the same set of samples. The optimization objective is to train a more accurate prediction model on the server by combining these features.

\textit{Definition 3.7.} (\textbf{Threat Model}). In the federated learning framework, attackers may launch attacks from both the client side and the server side. We assume that the server is curious but honest, meaning that the server can receive individual updates from different clients and use this information to infer the training data of each participant. However, external attackers may attempt to steal private information. These external attackers can infer from the SGD (Stochastic Gradient Descent) algorithm to perform active attacks, because the gradients computed by the SGD algorithm change depending on the importance of the training samples. If a training sample causes a large loss, the SGD algorithm adjusts the parameters to reduce the loss for that sample. If the training dataset does not contain this sample, the model’s loss change remains stable. Through repeated inference attacks, attackers can steal high-confidence private information.

\section{Methodology}
\noindent We consider a setup with a central server and $N$ local users. In this setup, each client $i \in N$ holds a private dataset $D_i=(d_1^i, d_2^i,..., d_R^i)$, where $R$ denotes the number of data points in the dataset. Let $D = D_1 \cup D_2 \cup...\cup D_N$ represent the union of all client datasets, with the assumption that these datasets are disjoint. Let $w$ be the model parameters. Our objective is to optimize the following function:
\begin{equation}
    \min_{w \in \mathbb{R}^d} \mathcal{L}(x) = \frac{1}{N} \sum_{i=1}^N \mathcal{L}_i(w) ,
\end{equation}
where $\mathcal{L}_i(w)$ represents the cross-entropy loss function for local training on client $i$, and the overall loss function $\mathcal{L}(w)$ is the weighted average of the losses across all clients.\par
Fig \ref{fig:overview} illustrates the overall architecture of GCFL. Each client first trains the model on its local dataset. In each iteration, the client selects a batch of training data using Poisson sampling and trains the model using the DP-SGD algorithm. Specifically, after computing the gradients of all model parameters, the client adds noise that satisfies the differential privacy requirements. The noisy gradients are then uploaded to the central server. The server is responsible for denoising and aggregating the received gradients. Finally, the server distributes the aggregated global gradients to all clients for updating their local models. This process iterates until the predefined model accuracy or training rounds are reached.

\subsection{Local Training Process}
During the local training phase, the objective is to ensure a certain level of training accuracy while mitigating the risk of information leakage from a single dataset $D_i$ when users share model updates. Based on the setting in DP-scaffold, our analysis focuses on record-level differential privacy in the combined dataset $D$. Specifically, for two neighboring datasets $D_i$ and $D_j$ (differing by a single data record, with all other parts identical), an attacker cannot determine the presence of a specific data record from the query results. All data remains stored locally during training, and clients cannot directly share their raw datasets. Each client $i$ maintains an identical DL model structure and uses its local dataset $D_i$ as input. The model output is denoted as $\hat{y}$. We adopt the cross-entropy loss function, expressed as:
\begin{equation}
    \mathcal{L}_i(w)=- \frac{1}{R} \sum_{j=1}^{R} y_j^ilog\hat{y}_j^i ,
\end{equation}
where $w$ represents the model parameters, $y$ is the true label, and $j$ indexes the $j$-th sample in the dataset.\par

During each iteration $t$, client $i \in N$ performs Poisson sampling to select a mini-batch $S$ of size $\lfloor sR \rfloor$ from its local dataset $D_i$, where $s$ is the sampling rate. Gradients $g_t(x)= \triangledown \mathcal{L}_t(w)$ are then computed for all data points in the batch $S$. To limit the influence of individual samples on the model updates within a predefined range, gradient clipping is applied based on a pre-determined clipping threshold $C_t$:
\begin{equation}
    g_t^{i-\text{clipped}}(w) = g_t(w) \cdot min(1,\frac{C_t}{{\| g_t^i(w) \|}_2}) .
\end{equation}
\par 

The purpose of the clipping threshold is to limit the update norm of all clients within $C_t$, ensuring that the impact of any single client's update on the final aggregated result, in terms of the L2 norm, does not exceed $C_t$. During the gradient descent process, an anomalous data point or one that has a significant impact on the loss function may produce a gradient or model update with a very large norm. Without clipping, this large-norm update could significantly affect the global model, potentially leaking sensitive information about the data. After clipping, the contribution of each sample to the gradient is restricted within the range $[0, C_t]$, thereby reducing the sensitivity to $2C/sR$. To further obscure the influence of sensitive data, Gaussian noise is added to the clipped gradients to further obscure the influence of sensitive data. The magnitude of the Gaussian noise is determined by the noise multiplier $\sigma$ and the clipping threshold, as detailed below:
\begin{equation}
    \hat{g}_t^i = \frac{1}{|S_t|}(\sum_{i \in S_t}g_t^{i-\text{clipped}}(w_i) + \mathcal{N}(0,{\sigma}^2{C_t}^2)) .
\end{equation}
\par The standard deviation $\sigma$ of the Gaussian noise controls the extent to which the algorithm satisfies $(\epsilon,\delta)-DP$. A larger $\sigma$ value increases the level of privacy protection but also amplifies the noise’s effect on the gradients, potentially degrading the model's performance.

\subsection{Server Aggregation Process}
\begin{algorithm}[!ht]
    \caption{Overall of GCFL}
    \label{alg:alg}
    \renewcommand{\algorithmicrequire}{\textbf{Input:}}
    \renewcommand{\algorithmicensure}{\textbf{Output:}}
    \begin{algorithmic}[1]
        \REQUIRE Training dataset $D_i=(d_1^i,d_2^i,\dots,d_R^i)$; hyperparameters: learning rate $\eta$, batch size, clipping threshold $C_t$, noise multiplier $\sigma$.
        \ENSURE The final trained model $w_t$
        
        \STATE Initialize $t = 1$, $w_0 = \text{Initial()}$
        \WHILE{$t < T$}
            \STATE User subsampling by the server:
            \STATE Sample $Cl_t \subset [N]$
            \STATE Server sends $w_{t-1}$ to user $i \in Cl_t$
            \FOR{user $i \in Cl_t$}
            \STATE Initialize model:
                \FOR{$k = 1,2,\dots,K$}
                    \STATE Data subsampling by user $i$:
                    \STATE Randomly sample a batch $S_i^k \subset D_i$ of size $\lfloor sR \rfloor$
                    \FOR{each sample $j \in S_i^k$}
                        \STATE Compute gradient:
                        \STATE $g_t^i(w) \leftarrow \nabla \mathcal{L}_t(w)$
                        \STATE Clip gradient:
                        \STATE $g_t^{i-\text{clipped}}(w) \leftarrow g_t(w) \cdot \min\left(1, \frac{C_t}{{\| g_t^i(w) \|}_2}\right)$
                    \ENDFOR
                    \STATE Add DP noise to local gradients:
                    \STATE $\hat{g}_t^i = \frac{1}{|S_i^k |}(\sum_{i \in S_i^k }g_t^{i-\text{clipped}}(w_i) + \mathcal{N}(0,{\sigma}^2{C_t}^2))$
                    \STATE User $i$ send $\hat{g}_t^i$ to server 
                    \ENDFOR
            \ENDFOR
        \STATE Server denoising and correction:
        \FOR{$\hat{g}_t^i \in \hat{g_t} $}
            \FOR{$\hat{g}_t^j \in \hat{g_t}$}
                \STATE Evaluate the impact of noise:
                \STATE $\cos \phi_{ij} = \frac{\hat{g}_t^i \cdot \hat{g}_t^j}{|\hat{g}_t^i||\hat{g}_t^j|}$
                \IF{ $\cos \phi_{ij}<$0}
                    \STATE $\hat{g}_t^{i-cor} = \hat{g}_t^i - \frac{\hat{g}_t^i \cdot \hat{g}_t^j}{\|\hat{g}_t^j\|_2}\hat{g}_t^j$
                \ENDIF
            \ENDFOR
        \ENDFOR
        \STATE $w_{t} \leftarrow w_{t-1} - \eta {\frac{1}{N}\sum_{i \in Cl_t}\hat{g}_t^{i-cor} }$
        \ENDWHILE
        
        \RETURN Final trained model $w_t$
    \end{algorithmic}
\end{algorithm}
% \vspace{-1 cm}
After clients hide their model information using differential privacy, the central server needs to aggregate the uploaded gradients $\hat{g_t^i}$. Unlike traditional methods that directly use weighted averages to combine gradients from different clients, we first perform denoising on these gradients. Our goal is to adjust the gradients in a way that promotes positive interactions among client gradients without making assumptions about the form of the model. However, completely eliminating the influence of Gaussian noise is unrealistic. Therefore, we need to identify gradients that severely impede the model's convergence in the correct direction and correct them.

To achieve this, in each iteration $t$, we randomly select a batch of clients and assume that their gradients in that iteration correctly point toward the model convergence direction. Subsequently, for all other client gradients, we determine whether they conflict with the correct gradient direction using the following equation:
\begin{equation}
    \cos \phi_{ij} = \frac{\hat{g}_t^i \cdot \hat{g}_t^j}{|\hat{g}_t^i||\hat{g}_t^j|} ,
\end{equation}
where $\hat{g}_t^i$ represents the model gradient uploaded by the $i$-th client in iteration $t$.\par

If the cosine similarity between the gradients of two different clients is negative, we consider that the gradient has deviated from the correct direction due to the added noise and needs correction. For the gradient $\hat{g}_t^i$ that conflicts with the correct gradient $\hat{g}_t^j$, we apply the following correction method:
\begin{equation}
    \hat{g}_t^{i-cor} = \hat{g}_t^i - \frac{\hat{g}_t^i \cdot \hat{g}_t^j}{\|\hat{g}_t^j\|_2}\hat{g}_t^j .
\end{equation}
This operation projects the gradient $\hat{g}_t^i$ onto the normal plane of gradient $\hat{g}_t^j$. Intuitively, this is equivalent to removing the conflicting component of the gradient, thereby reducing noise interference. From a vector analysis perspective, the gradients uploaded by clients can be viewed as the superposition of the original gradient and Gaussian noise. Since clients are training on the same task, their original gradient directions usually do not have significant discrepancies; thus, the projection operation does not significantly affect the original gradients. As for the random noise, because it is difficult for it to align with the complex gradient directions in high-dimensional space, it may hinder model updates or even oppose the original gradient direction. This portion of the noise will be weakened or eliminated by the projection operation, and only a small amount of noise close to the original gradient direction will be retained. The overall computational complexity of gradient correction is $O(M \cdot (N-M)\cdot D + K \cdot D)$, where M represents the number of clients selected in each iteration, N is the total number of clients participating in the federated learning process

The server calibrates the updated gradients received from the remaining clients using a randomly selected subset of client gradients. This process continues until the gradients from all clients are adjusted towards a similar direction, effectively mitigating the detrimental effects of high-magnitude noise. Subsequently, the server aggregates the gradients from the different clients using a weighted aggregation scheme. The weights for this aggregation process are determined based on the number of training samples available at each client during the current iteration. After the gradients from all clients have been denoised and aggregated through weighted averaging, the central server broadcasts the resulting global model update to all clients. Each client then uses this update to refine its local model via backpropagation. This iterative process continues until a predetermined number of training rounds is completed or the desired model accuracy is achieved. The overall GCFL procedure is shown in Algorithm \ref{alg:alg}.

\section{Experimental Evaluation}
\noindent This section presents the empirical evaluation of GCFL against baseline models on several datasets. Section \ref{s:Experimental Setting} details the datasets used, the baseline models employed for comparison, and the evaluation metrics. The experimental results and analysis are presented in Section \ref{s:Main Results}.
\subsection{Experimental Setting}
\label{s:Experimental Setting}
% \noindent
\begin{table}[htpb]
\caption{Experimental parameter settings.}
\label{parameter}
\begin{tabular}{|l|l|}
\hline
Parameter                        & Value        \\ \hline
Number of edge servers $N$         & 2            \\ \hline
Total Number of epochs $T$         & \{30,35,80\} \\ \hline
Learning Rate $\eta$                    & \{0.001,0.002,0.005\}        \\ \hline
Batch size for training          & \{32,48\}          \\ \hline
Batch size for testing           & 1024         \\ \hline
Differential privacy noise multiplier $\sigma$  & 0.8          \\ \hline
Maximum clipping threshold per sample $C_t$ \hspace{0.5cm} & 1.5          \\ \hline
Privacy leakage probability $\delta$      & 1e-5         \\ \hline
\end{tabular}
\end{table}

\begin{table*}[htpb]
% \vspace{+0.15cm}
\scriptsize
\centering
\caption{Comparison of accuracy, recall and F1 score on three image datasets ($\epsilon$=2, epoch=60). Best results in bold. Isolated denotes local-model training.}
\vspace{+0.15cm}
\label{table_1}
\resizebox{0.8\textwidth}{!}{
\begin{tabular}{lccccccccc}
\hline
\multirow{2}{*}{Models} & \multicolumn{3}{c}{COVID-19}           & \multicolumn{3}{c}{MNIST}                        & \multicolumn{3}{c}{CIFAR-10}                     \\ \cline{2-10} 
                        & ACC            & REC       & F1        & ACC            & REC            & F1             & ACC            & REC            & F1             \\ \hline
Isolate                 & 64.01          & 54.28     & 53.68     & 79.14          & 78.60          & 77.78          & 65.58          & 65.58          & 65.52          \\
DP-FedAvg               & 69.10           & 60.85     & 60.51     & 85.50          & 85.26          & 85.12          & 75.61          & 73.61          & 75.75          \\
DP-FedProx              & 68.28          & 58.38     & 58.06     & 86.28          & 86.07          & 85.97          & 75.55          & 75.55          & 75.53          \\
DP-Scaffold             & 69.33          & 61.46     & 61.12     & 87.13          & 86.96          & 86.92          & 74.68          & 74.68          & 74.75          \\
DP-FedExP               & 70.21          & 63.79     & 63.55     & 89.37          & 88.94          & 89.22          & 73.93          & 73.58          & 73.43          \\ \hline
GCFL                    & \textbf{71.34} & \textbf{65.17} & \textbf{67.29} & \textbf{91.11} & \textbf{91.02} & \textbf{91.03} & \textbf{76.83} & \textbf{76.60} & \textbf{76.58} \\ \hline
\end{tabular}
}
\end{table*}

\begin{table*}[htpb]
\scriptsize
\centering
\caption{Comparison with baseline models under different privacy budgets on the MNIST dataset. The best performance results from this experiment are highlighted in bold.}
\vspace{+0.3cm}
\label{table_2}
\resizebox{0.8\textwidth}{!}{
\begin{tabular}{lccccccccc}
\hline
\multirow{2}{*}{Models} & \multicolumn{3}{c}{$\epsilon=2$}                 & \multicolumn{3}{c}{$\epsilon=3$}                 & \multicolumn{3}{c}{$\epsilon=4$}                 \\ \cline{2-10} 
                        & ACC            & REC            & F1             & ACC            & REC            & F1             & ACC            & REC            & F1             \\ \hline
DP-FedAvg               & 85.50          & 85.26          & 85.12          & 88.73          & 88.62          & 88.59          & 90.51          & 90.43          & 90.41          \\
DP-FedProx              & 86.28          & 86.07          & 85.97          & 90.57          & 90.49          & 90.47          & 92.18          & 92.13          & 92.12          \\
DP-Scaffold             & 87.13          & 86.96          & 86.92          & 90.98          & 90.82          & 90.90          & 92.13          & 92.09          & 92.09          \\
DP-FedExP               & 89.07          & 88.94          & 89.22          & 91.57          & 91.37          & 91.47          & 92.33          & 92.28          & 92.31          \\ \hline
GCFL                    & \textbf{91.11} & \textbf{91.02} & \textbf{91.03} & \textbf{92.76} & \textbf{92.43} & \textbf{92.70} & \textbf{93.15} & \textbf{93.13} & \textbf{93.09} \\ \hline
\end{tabular}
}
\end{table*}

\textbf{Datasets.} The performance of the proposed GCFL framework is evaluated using three benchmark datasets: COVID-19 Radiography, MNIST, and CIFAR-10. These datasets represent a range of task scenarios, allowing us to assess the framework’s effectiveness in noisy edge environments.

The COVID-19 Radiography dataset includes 21,165 chest X-ray images, consisting of 10,192 normal images, 6,012 with lung opacities, and 1,345 with viral pneumonia. All images have a resolution of 299×299 pixels, with a training set of 17,993 images and a test set of 3,172 images.

The MNIST dataset comprises 70,000 images of handwritten digits, each in grayscale with a 28×28 pixel resolution. The dataset is partitioned into 60,000 training samples and 10,000 test samples, with 7,000 images per digit class.

The CIFAR-10 dataset contains 60,000 color images, equally distributed among 10 categories, with 5,000 images per category. All images have dimensions of 32×32 pixels, with 50,000 images allocated for training and 10,000 for testing.

The non-IID dataset is constructed by evenly splitting the original MNIST training set of 60,000 images into two contiguous halves. The first 30,000 samples (labels 0–4) are assigned to client 1, and the remaining 30,000 samples (labels 5–9) to client 2. This preserves MNIST’s image resolution and overall class proportions while introducing non‑IID heterogeneity across the two clients.
\par

\textbf{Baselines.} To evaluate the effectiveness of our framework, we conducted a comparative analysis with the following baseline methods.
\begin{itemize}
    \item \textbf{DP-FedAvg} \cite{mcmahan2017communication} This is the most classic federated learning algorithm. The server updates the global model by taking a weighted average of the model parameters uploaded by the clients, making it both simple and efficient.
    \item \textbf{DP-FedProx} \cite{li2020federated} This approach seeks to mitigate the performance decline resulting from data heterogeneity and inconsistency. By incorporating a regularization term into the local client optimization objective, it constrains the disparity between local updates and the global model, thereby avoiding significant deviations from the global model during each local training phase.
    \item \textbf{DP-Scaffold} \cite{karimireddy2019scaffold} The goal of this approach is to reduce the bias between local updates on different clients. The server maintains a set of control variates, and when clients update their local models, these control variates are taken into account to eliminate the bias in the local model updates.
    \item \textbf{DP-FedExP} \cite{jhunjhunwala2023fedexpspeedingfederatedaveraging} This method adaptively adjusts the server's step size based on the dynamically changing pseudo-gradient during the FL process. It builds on the relationship between FedAvg and the projection onto convex sets (POCS) algorithm, providing an extrapolation mechanism that accelerates convergence, especially in over-parameterized convex problems.
\end{itemize}

\par \textbf{Evaluation Metrics.} To ensure consistency and facilitate fair comparison with existing methods, we employed identical hyperparameter settings and evaluation metrics across all experiments.  Specifically, the number of clients was fixed at 2, the differential privacy parameter $\delta$ was initialized to $10^{-5}$, the learning rate $\eta$ was tuned amongst $\{0.001, 0.002, 0.005\}$, and the gradient clipping threshold was set to 1.5.  The remaining hyperparameters are detailed in Table \ref{parameter}.  Comparative experiments were conducted on a simplified network architecture comprising only convolutional and fully connected layers.  Following standard practice in the field, we evaluated model performance using accuracy, recall, and F1-score as primary metrics.  The reported results represent the average performance across three independent runs conducted on an NVIDIA RTX 4070 Super GPU.  Accuracy is defined as the ratio of the number of correctly predicted samples to the total number of samples. Recall measures the model’s ability to identify all positive class samples, while the F1-score \cite{chicco2020advantages} is an important metric for evaluating the overall performance of a classifier, especially in the case of imbalanced classes. 
% The specific formula for calculating the F1-score is as follows:
% \begin{equation}
%     F1-score = \frac{2 \cdot TP}{2 \cdot TP+FP+FN}
% \end{equation}

\subsection{Main Results}
\label{s:Main Results}

\begin{figure*}[ht]
    \vspace{+0.3cm}
    \centering
    \includegraphics[width=1\linewidth]{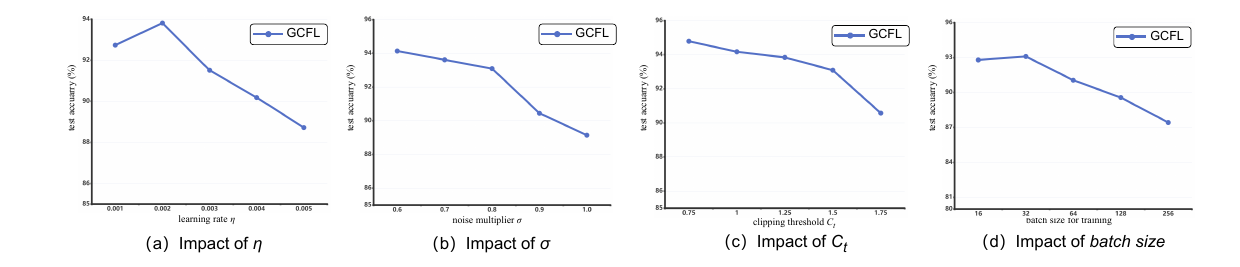}
    \centering
    \caption{The impact of different parameters on the test accuracy in MNIST dataset.
}
    \label{fig:parameters}
    \vspace{+0.2cm}
\end{figure*}

\noindent
% In this section, we compare the performance of our proposed framework with existing methods using three benchmark datasets. The evaluation was conducted under the same privacy budget, and the results are summarized in Table \ref{table_1}. Our framework outperforms other baseline methods across all datasets, demonstrating that the gradient correction mechanism can effectively identify and denoise perturbed gradients that disrupt correct model updates while preserving crucial information from the original gradients. Notably, on the MNIST dataset, our framework achieves significant improvements compared to all baselines. This may be attributed to the relative simplicity of the MNIST dataset, allowing the model to quickly capture feature relationships and enabling our framework to efficiently aggregate updates from different clients. However, on the more complex COVID-19 and CIFAR-10 datasets, although our framework still outperforms other baselines, the performance gains are less pronounced than on MNIST. This could be partly due to the simplified convolutional network architecture employed in this experiment, and partly because, on complex datasets, the model struggles to find optimal solutions along the gradient direction in the early stages of training. Projecting excessive noisy gradients might hinder model performance and convergence speed.\par

In this section, we evaluate the performance of our proposed framework against existing methods using three benchmark datasets. All evaluations were conducted under the same privacy budget, with the results summarized in Table \ref{table_1}. Across all datasets, our framework consistently outperforms the baseline methods, demonstrating the efficacy of the gradient correction mechanism in identifying and denoising perturbed gradients that impair proper model updates, while retaining critical information from the original gradients. Notably, on the MNIST dataset, our framework achieves significant improvements over all baselines. This may be attributed to the dataset's relative simplicity, which allows the model to quickly learn feature relationships and enables our framework to more effectively aggregate updates from different clients. In contrast, on the more complex COVID-19 and CIFAR-10 datasets, although our framework still surpasses the baselines, the performance gains are comparatively less pronounced. This may be due to the use of a simplified convolutional network architecture in our experiments, and also because on more complex datasets, the model finds it more difficult to follow optimal gradient directions during the early training stages. In such cases, projecting excessively noisy gradients can impede performance and slow convergence. \par

We evaluated the performance of our framework in different privacy budget settings by comparing it with leading competitive frameworks. Performance in the MNIST dataset was tested under various privacy budgets, and the results are summarized in Table \ref{table_2}. Remarkably, our framework consistently exhibits the best performance under all conditions, achieving a significant balance between accuracy and privacy preservation. This both affirms the framework's effectiveness and showcases its robustness under different privacy budget conditions, making it appropriate for applications requiring stringent privacy protection. Our framework demonstrates a substantial advantage over other frameworks under strict privacy constraints ($\epsilon$=2). This is because frameworks such as DP-FedAvg lack additional mechanisms to handle the noise introduced by differential privacy. Consequently, under high noise conditions, the model's update direction is misled by erroneous gradients, hindering convergence. GCFL effectively addresses this issue.
\par

\begin{figure*}[htpb]
    \centering
    % \vspace{-0.4cm}
    \includegraphics[width=1\linewidth]{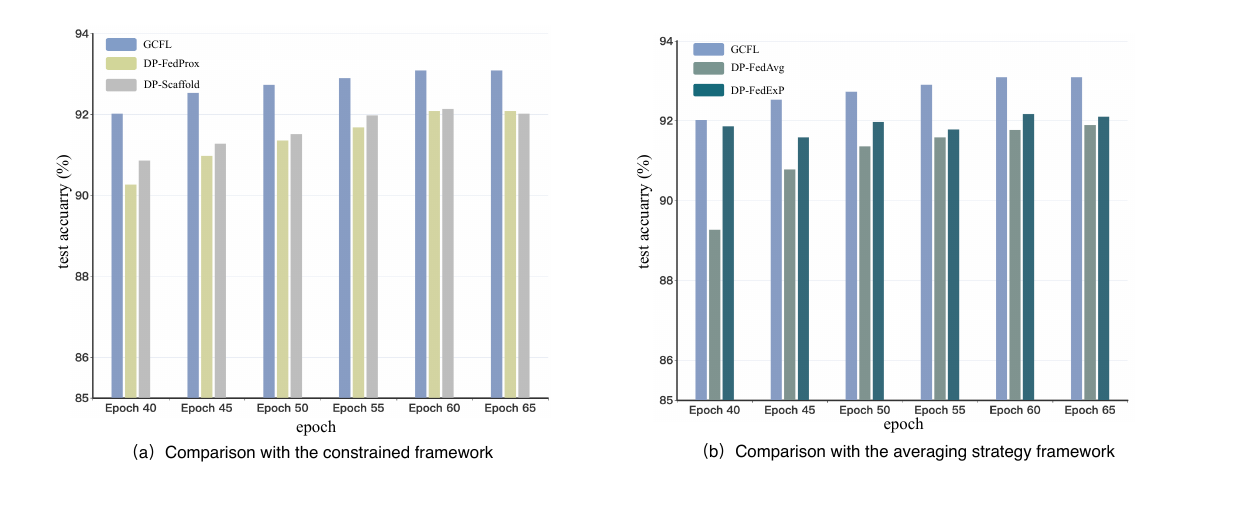}
    % \vspace{-0.8cm}
    \caption{Comparison of different frameworks and epochs on the MNIST dataset.}
    \label{fig:epoch}
    \vspace{+0.5cm}
\end{figure*}

\begin{figure*}[htpb]
    \centering
    % \vspace{-0.4cm}
    \includegraphics[width=1\linewidth]{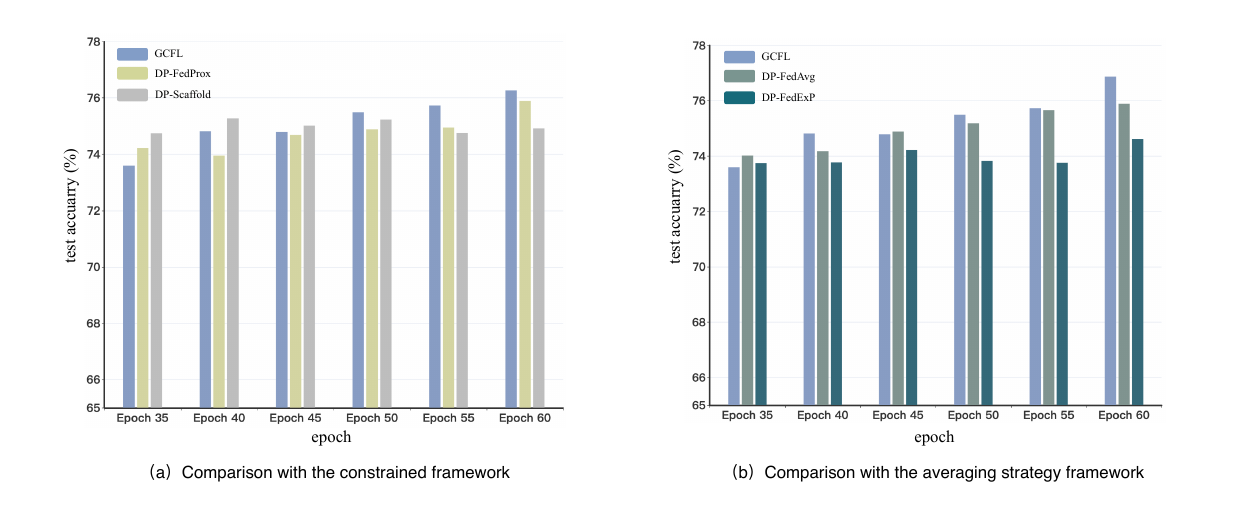}
    % \vspace{-0.8cm}
    \caption{Comparison of different frameworks and epochs on the CIFAR-10 dataset.}
    \label{fig:epoch1}
    % \vspace{+0.3cm}
\end{figure*}

\begin{figure}[htpb]
    \centering
    % \vspace{-0.4cm}
    \includegraphics[width=0.9\linewidth]{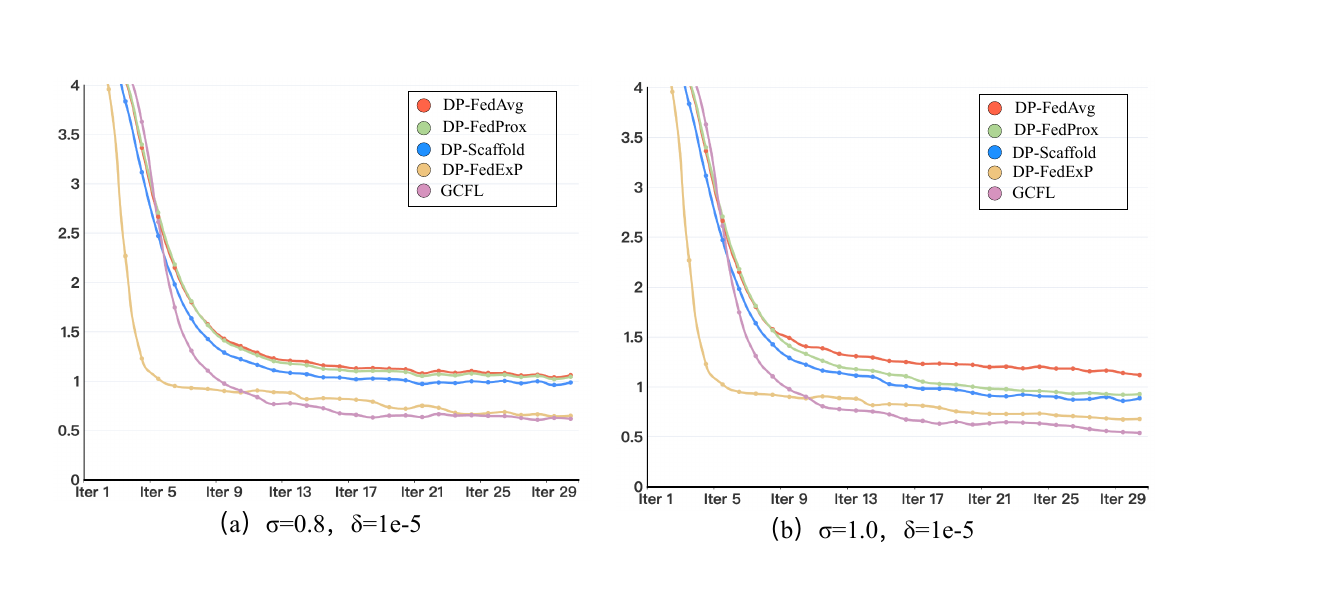}
    % \vspace{-0.8cm}
    \caption{Impact of differential privacy noise levels on the training loss convergence performance.}
    \label{fig:loss}
    % \vspace{+0.5cm}
\end{figure}

Figure \ref{fig:parameters} illustrates the impact of various hyperparameters on the GCFL framework.  A peak accuracy of 93.81\% is achieved with a learning rate $\eta$ of 0.002.  For $\eta > 0.002$, the accuracy begins to decline, likely due to the large learning rate hindering model convergence.  Figures \ref{fig:parameters}(b) and (c) demonstrate the robustness of GCFL to variations in $\sigma$ and $C_t$, parameters that influence the noise added for differential privacy.  The performance of GCFL does not fluctuate dramatically with changes in these parameters.  It is worth noting that excessively large values for $\sigma$ and $C_t$ do not yield further benefits. This is because very large values can lead to the model correcting gradients that deviate significantly from the convergence direction, resulting in corrected gradients approaching zero and contributing minimally to model convergence. As shown in Figure \ref{fig:parameters}(d), GCFL achieves optimal performance with a batch size of 32.\par

We further compared the test accuracy of GCFL against baseline models on both the MNIST and CIFAR-10 datasets across different epochs.  The baseline models were categorized into two groups: Constrained Framework, encompassing DP-Prox and DP-Scaffold, which incorporate additional constraints or global correction techniques; and Averaging Strategy Framework, consisting of DP-FedAvg and DP-FedExP.  The comparative results are presented in Figures \ref{fig:epoch} and \ref{fig:epoch1}.  As shown, the performance of the Constrained Framework models (DP-Prox and DP-Scaffold) is similar on MNIST, but diverges significantly on CIFAR-10. While DP-FedAvg and DP-FedExP demonstrate faster early convergence on CIFAR-10 than GCFL, suggesting a more rapid initial aggregation and accuracy improvement, their final performance falls short of GCFL. In contrast, GCFL exhibits superior stability and robustness across both datasets. This indicates that GCFL more effectively coordinates the learning process across participants, maintaining model consistency and performance in the face of varying data distributions and complexities. This stability likely stems from GCFL striking a better balance between global consensus and local model optimization, leading to superior performance in diverse data environments.
\par 

\begin{table}[ht]
\centering
\caption{Comparison with baseline models on non-iid datasets.
% The best performance results from this experiment are highlighted in bold.
}
% \vspace{+0.5cm}
\label{non-iid}
\begin{tabular}{lccc}
\hline
Models      & ACC                       & REC                       & F1                        \\ \hline
DP-FedAvg   & 84.53                     & 84.31                     & 84.12                     \\
DP-FedProx  & 86.95                     & 86.77                     & 86.57                     \\
DP-Scaffold & 86.98                     & 86.19                     & 86.84                     \\
DP-FedExP   & 87.69                     & 87.35                     & 87.34                     \\ \hline
GCFL        & \multicolumn{1}{l}{\textbf{88.98}} & \multicolumn{1}{l}{\textbf{88.80}} & \multicolumn{1}{l}{\textbf{88.82}} \\ \hline
\end{tabular}
\end{table}

\begin{table}[htpb]
% \scriptsize
\centering
\caption{Comparison of speed (samples per second) between the baseline model and GCFL. 
% The best performance results from this experiment are highlighted in bold.
}
% \vspace{+0.5cm}
\label{speed}
\begin{tabular}{lcccc}
\hline
\multirow{2}{*}{Model} & \multicolumn{2}{c}{MNIST}           & \multicolumn{2}{c}{CIFAR-10}        \\ \cline{2-5} 
                       & Train            & Test             & Train            & Test             \\ \hline
DP-FedAvg              & 8109.43          & 11447.12         & 6165.22          & 9184.04          \\
DP-FedProx             & 7669.80           & 11765.76         & 5731.56          & 9021.14          \\
DP-Scaffold            & 7801.85          & 11787.53         & 5913.84          & 8725.34          \\
DP-FedExP              & \textbf{8221.38} & \textbf{11847.57} & \textbf{6310.38} & \textbf{9359.29} \\ \hline
GCFL                   & 7598.38          & 11347.57         & 5639.38          & 8772.95          \\ \hline
\end{tabular}
\end{table}

% To further investigate model loss behavior under high noise regimes, we recorded the training loss of different models with $\sigma$ = 0.8 and $\sigma$ = 1.0. The results are presented in Figure \ref{fig:loss}. As shown, DP-FedExP exhibits superior convergence speed in the initial training phase. This can be attributed to the inherent design of DP-FedExP, which combines the advantages of DP-FedAvg and Projection Onto Convex Sets (POCS). By incorporating an extrapolation mechanism, DP-FedExP utilizes current gradient information more effectively, accelerating the parameter update process. This extrapolation mechanism is particularly effective in over-parameterized convex optimization problems, significantly reducing the number of iterations required for convergence and thus leading to a rapid decrease in model loss during the early stages. Although DP-FedExP converges faster initially, GCFL demonstrates more robust performance throughout the entire training process, effectively exploring the global optimal solution space. By coordinating model updates across participants, GCFL ensures global model consistency, thereby improving the overall generalization ability and stability of the model and avoiding local optima.\par

As illustrated in Figure 6(a), under moderate noise ($\sigma=0.8$), all methods initially exhibit rapid loss reduction. DP-FedExP, utilizing its POCS extrapolation mechanism, achieves the fastest initial convergence, reducing the loss to approximately 0.35 within the first 30 rounds. However, its convergence rate diminishes notably between rounds 30 and 50, accompanied by minor loss oscillations thereafter. DP-FedProx and DP-Scaffold also display fluctuations in later training stages, suggesting that conventional regularization or global correction mechanisms offer limited mitigation of gradient deviations under these conditions. In contrast, GCFL maintains a consistently smooth loss decrease throughout training, ultimately converging to the lowest loss value. This result confirms that the gradient projection-based correction mechanism effectively filters noise components, promotes stable convergence, and enhances final model performance.

Furthermore, under stricter privacy constraints ($\sigma=1.0$), as shown in Figure 6(b), the initial convergence of all baseline methods is significantly impeded. DP-FedExP's loss decreases only to about 0.65 within the first 20 rounds before plateauing. DP-FedAvg exhibits substantial fluctuations under high noise, with its final converged loss remaining above 0.6, indicating its simple averaging strategy is insufficient to counteract strong noise interference. Although DP-FedProx and DP-Scaffold demonstrate a marginal advantage during the intermediate stages, they also suffer from oscillations and an increase in loss after round 40. In sharp contrast, GCFL sustains a continuous and stable loss reduction even in this high-noise setting. After round 50, its loss drops below 0.25, outperforming the second-best method by achieving a final loss that is approximately 3\%–5\% lower. This underscores the effectiveness of the proposed gradient correction mechanism in eliminating erroneous gradient components while preserving informative ones, even within challenging high-noise environments.

To validate the effectiveness of the proposed GCFL framework in addressing the local drift issue under data heterogeneity, we conducted experiments on non-IID datasets. The experimental results in Table \ref{non-iid} demonstrate that GCFL consistently outperforms all baseline models across all three evaluation metrics. Specifically, when compared to the simplest baseline model, DP-FedAvg, which relies on weighted gradient averaging, GCFL exhibits significant improvements: an increase of 4.45\% in accuracy, 4.49\% in recall, and 4.70\% in F1 score. These findings indicate that GCFL effectively mitigates the local drift issue prevalent in federated learning with non-IID data by more effectively aligning client gradients to a consistent global direction.

In summary, under the same privacy budget, GCFL outperforms all comparison methods on three datasets. At the same time, GCFL maintains the highest accuracy across all privacy budgets, with the most notable improvement observed under strict privacy conditions. This is due to the gradient correction mechanism in GCFL, which effectively counteracts the noise distortion. In contrast, other methods struggle to converge due to excessive noise. In terms of convergence, comparison methods converge quickly in the early stages of training. However, they lag behind our model in the later stages. GCFL demonstrates better stability and final performance over a larger number of iterations, effectively avoiding local optima, continuing to explore the global optimal solution, and coping with local drift issues. Finally, we examined the computational overhead of GCFL, and the results show that with the introduction of a server-side projection correction mechanism, the performance overhead is minimal, resulting in a reasonable trade-off for a significant accuracy improvement.

Finally, we evaluated both the training and inference speeds of GCFL against baseline models. Experiments were conducted with a batch size of 48 and a learning rate of 0.005. As shown in Table \ref{speed}, even with the added gradient projection mechanism on the server, GCFL's training and inference speeds remain comparable to DP-FedProx and DP-Scaffold. DP-FedAvg and the built-upon DP-FedExP achieve suboptimal and optimal performance, respectively. We attribute this to FedAvg's simple weighted averaging of client gradients. The modest trade-off in training and inference speed incurred by GCFL is justified by the gains in predictive performance. \par

\section{Conclusions}
\noindent \noindent
This paper presents a novel differentially private federated learning framework designed to address the crucial challenges of privacy preservation and model performance enhancement in Cyber-Physical-Social Systems applications. By combining the strengths of differential privacy and federated learning, our framework mitigates the negative impact of noise during model aggregation while guaranteeing data privacy.  The innovative server-side gradient correction mechanism, based on gradient projection, plays a vital role in overcoming the challenges posed by noisy updates and local optima.  This mechanism ensures data privacy for individual clients and simultaneously boosts the global model's convergence speed and accuracy.

We are considering enhancing the framework's generalizability, for instance, by adapting the loss function to further improve its performance on non-IID datasets.  Furthermore, because our framework requires traversing gradients from different clients on the server side, the training speed per iteration is slightly slower than mainstream models. We plan to address this performance bottleneck in future work.
\par

\section*{Acknowledgment}
This work was supported in part by the National Natural Science Foundation of China under Grant 92267104 and Grant 62372242, Jiangsu Provincial Major Project on Basic Research of Cutting-edge and Leading Technologies, under grant no. BK20232032, and the Natural Science Foundation of Jiangsu Province under Grant BK20240692.

\bibliographystyle{IEEEtran}
\bibliography{refer}

% \vspace{-1.5cm}
\begin{IEEEbiography}[{\includegraphics[width=1in,height=1.25in,clip,keepaspectratio]{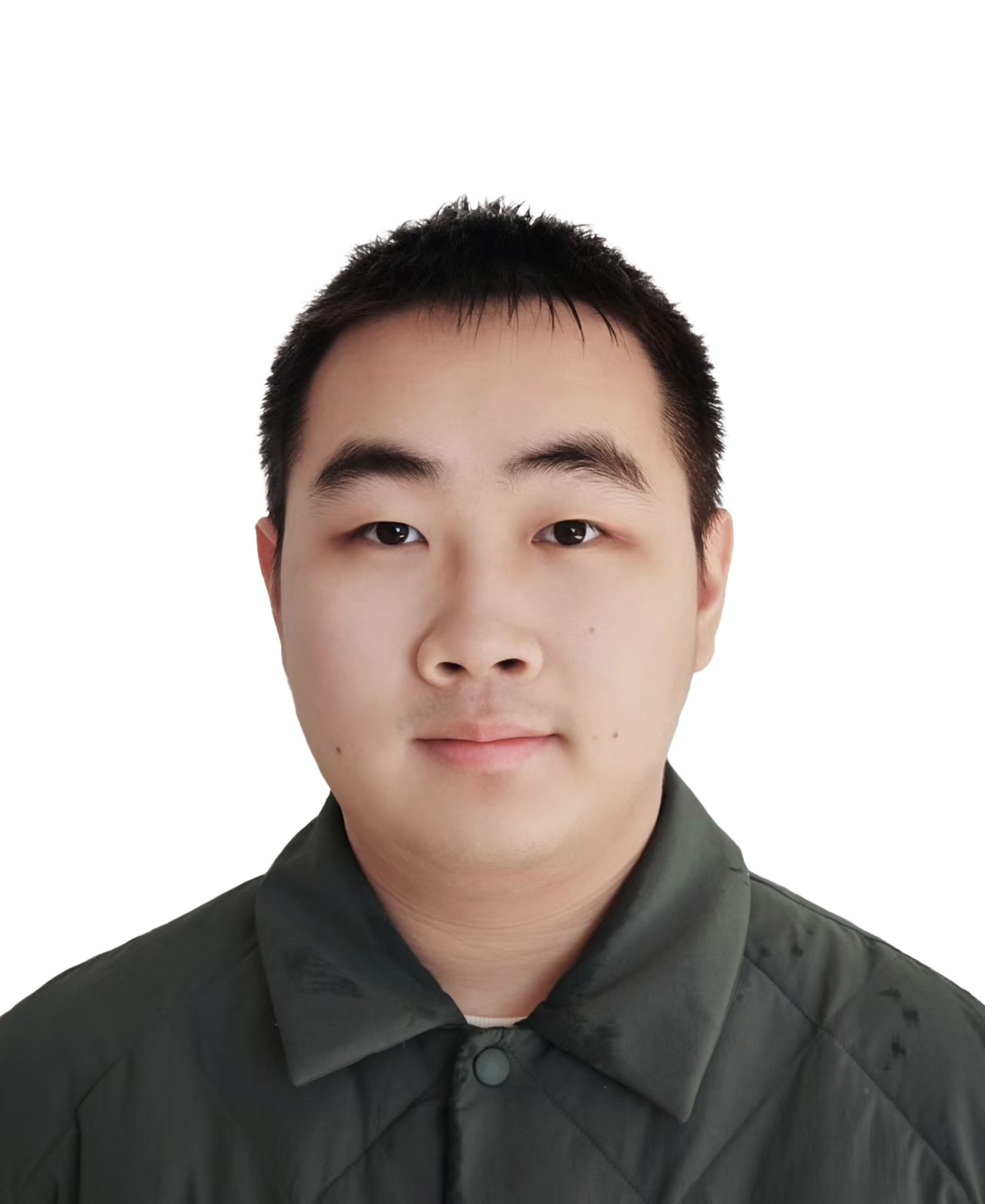}}]{Jiayi Wan} is currently pursuing a B.S. degree in Software Engineering at the School of Software Engineering, Nanjing University of Information Science and Technology. His research interests focus on federated learning, differential privacy, and information extraction. Contact him at wanjiayi117@gmail.com.\end{IEEEbiography}

% \begin{IEEEbiography}[{\includegraphics[width=1in,height=1.25in,clip,keepaspectratio]{pictures/GuangMing Cui.jpg}}]{Guangming Cui} received his Master's degree from Anhui University, China, in 2018 and his PhD degree from Swinburne University of Technology, Australia, in 2022, in computer science. Currently, he is an associate professor at Nanjing University of Information Science \& Technology, China.  His research interests include edge computing, service computing, mobile computing and software engineering. Contact him at gcui@nuist.edu.cn\end{IEEEbiography}
% % \vspace{+1.2cm}

\begin{IEEEbiography}[{\includegraphics[width=1in,height=1.25in,clip,keepaspectratio]{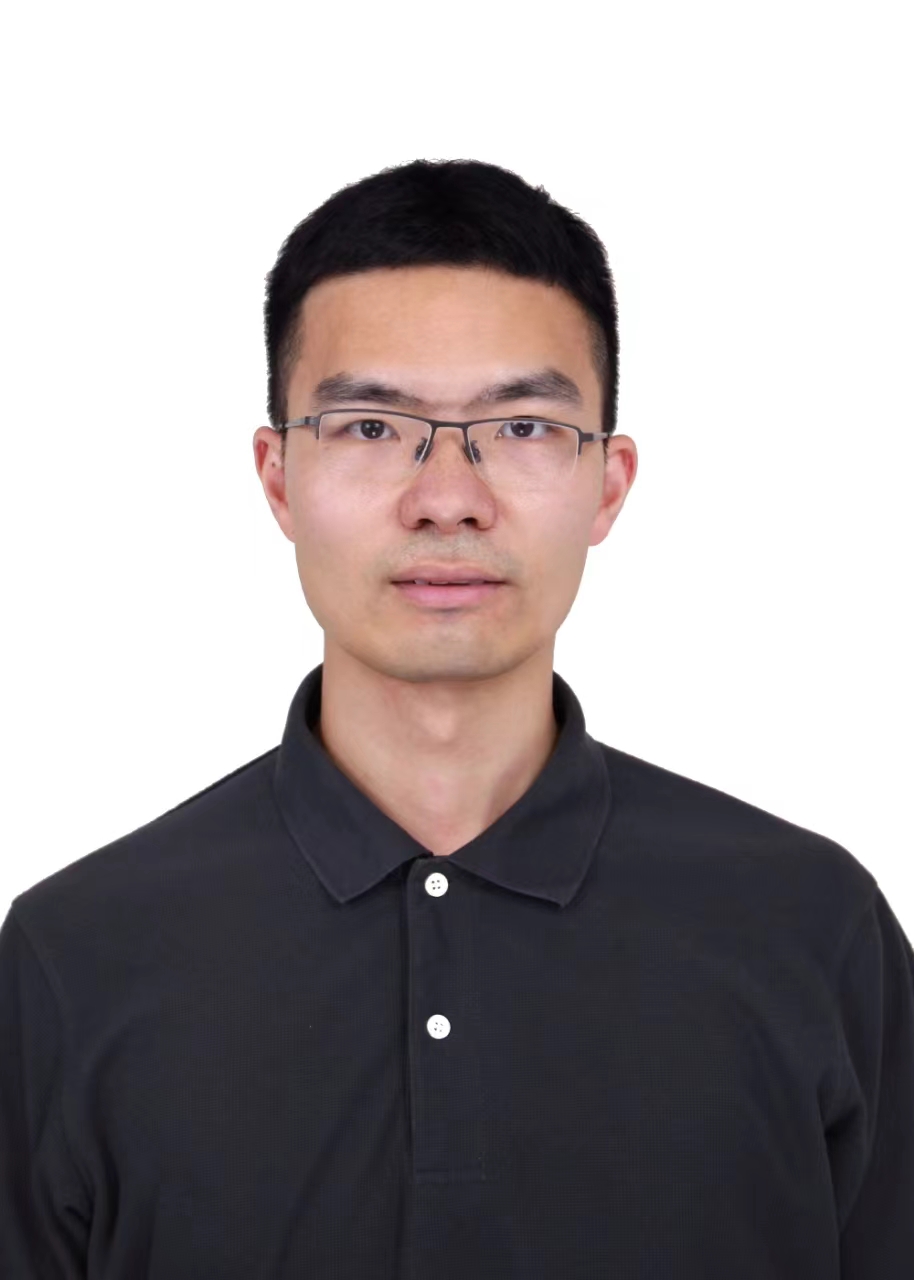}}]{Xiang Zhu} received a B.E. degree from Tsinghua University, Beijing, China, in 2010 and an M.E. degree from National University of Defense Technology, Hunan, China, in 2013. Currently, he is Ph.D. student at the College of Computer, National University of Defense Technology, Hunan, China. His research interests include data mining and big data analysis. Xiang Zhu has published in prestigious journals and
conferences, such as ICC, TREC, and ICPADS. Contact him at zhuxiang@nudt.edu.cn\end{IEEEbiography}

\begin{IEEEbiography}[{\includegraphics[width=1in,height=1.25in,clip,keepaspectratio]{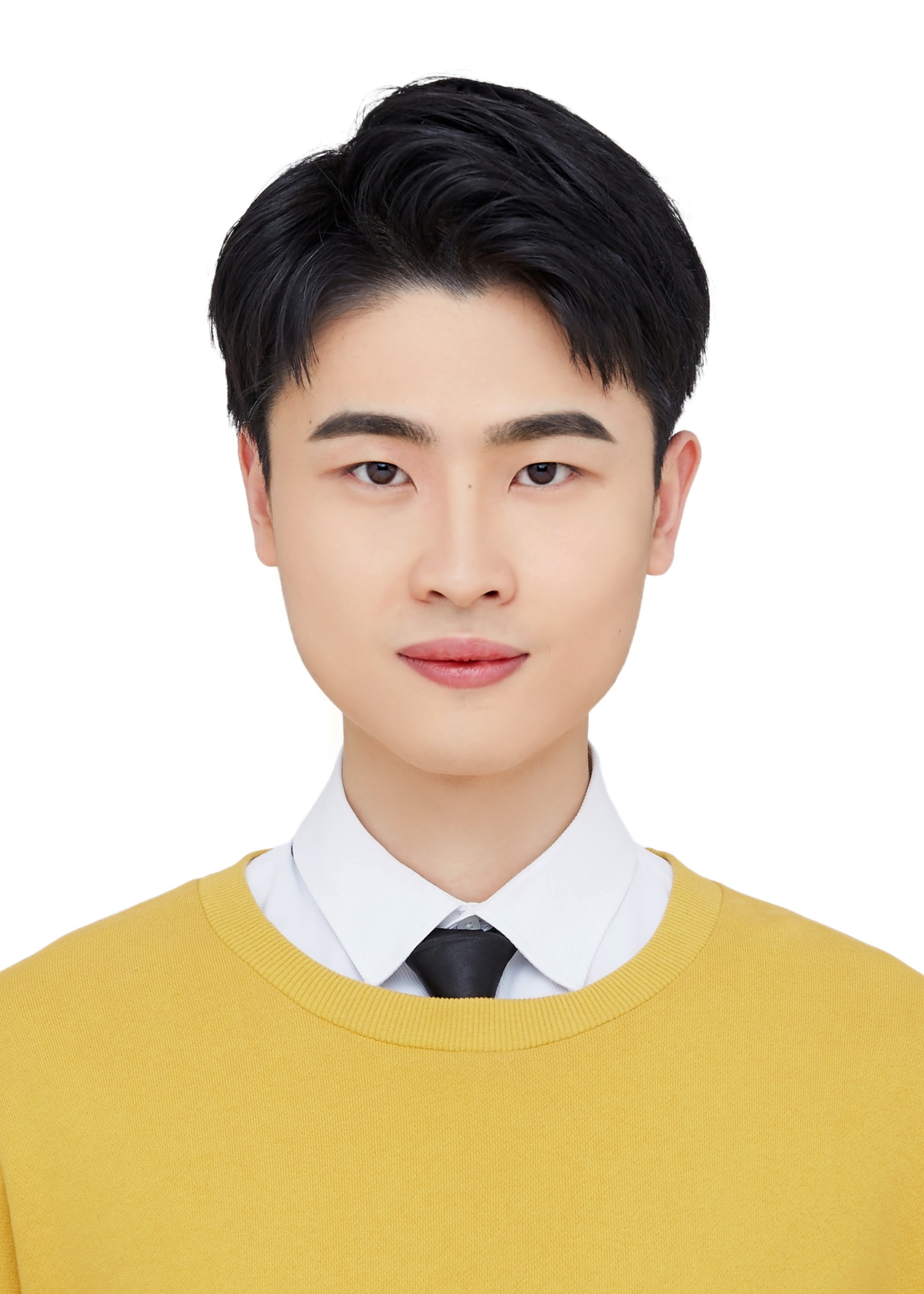}}]{Fanzhen Liu} earned his PhD in Computer Science from Macquarie University, Sydney, Australia, in 2024, where he currently serves as a Postdoctoral Research Fellow. He is also a Visiting Scientist at CSIRO’s Data61, Australia. His research interests include graph mining, anomaly detection, trustworthy machine learning, and social network analysis. Dr. Liu has published in prestigious journals and conferences, such as VLDB J., IEEE TKDE, IEEE TNNLS, KDD, ICDM, and IJCAI. Contact him at Fanzhen.Liu@data61.csiro.au\end{IEEEbiography}
% \vspace{+5cm}

\begin{IEEEbiography}[{\includegraphics[width=1in,height=1.25in,clip,keepaspectratio]{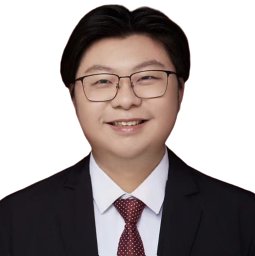}}]{Wei Fan} is currently working as a Postdoctoral Researcher in the Medical Sciences Division at the University of Oxford, UK. His research focuses on data-centric AI, time series modeling, and spatial-temporal data mining. Wei earned his Ph.D. degree in Computer Science from the University of Central Florida. Wei has published over 20 papers in ICLR, NeurIPS, AAAI, IJCAI, TKDE, etc. Contact him at frankfanwei@outlook.com.\end{IEEEbiography}
% \vspace{-15cm}

\begin{IEEEbiography}[{\includegraphics[width=1in,height=1.25in,clip,keepaspectratio]{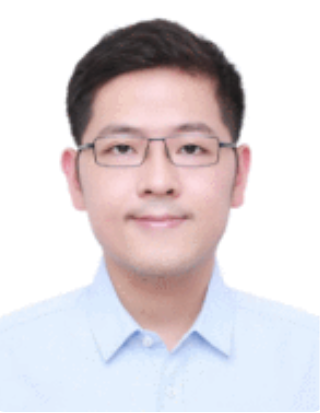}}]{Xiaolong Xu} (Senior Member, IEEE) was a Research Scholar with Michigan State University, East Lansing, MI, USA, from April 2017 to May 2018. He is currently a Full Professor with the School of Software, Nanjing University of Information Science and Technology, Nanjing. He has authored or co-authored more than 60 peer-reviewed articles in international journals and conferences including TKDE, ITPDS, TSC, TFS, TITS, and ICSOC. His research interests include edge computing, the IoT, cloud computing, big data, and service computing.\end{IEEEbiography}

\end{document}